\documentclass[11pt]{article}
\usepackage{times} 
\usepackage{url} 
\usepackage{graphicx} 
\usepackage[margin=1in]{geometry} 
\usepackage{authblk} 
\usepackage{amsmath} 
\usepackage[table]{xcolor}  
\usepackage{colortbl}       
\definecolor{lightblue}{RGB}{200, 230, 255}

\usepackage{amsfonts}
\usepackage{tabularx,ragged2e,booktabs} 
\usepackage{titlesec}
\usepackage{amssymb}
\usepackage{comment}
\usepackage{subcaption}
\usepackage{authblk}
\usepackage{xcolor} 
\usepackage{hyperref}
\usepackage{amssymb}
\usepackage{dsfont}
\usepackage{float}
\usepackage[algo2e,ruled,vlined]{algorithm2e}
\usepackage{multirow}
\usepackage[normalem]{ulem}
\usepackage{graphicx,lipsum,wrapfig}
\usepackage{cleveref}
\usepackage[flushleft]{threeparttable}

\usepackage{booktabs,caption}
\usepackage[sorting=none,style=ieee,citestyle=numeric-comp, backend=biber]{biblatex}
\bibliography{bib}

\usepackage{wrapfig}
\usepackage{color}

\newcommand{\emdash}{---}

\begin{document}


\author{
    Michal Golovanevsky$^{\dagger 1}$, Pranav Mahableshwarkar$^{\dagger 1}$, Carsten Eickhoff$^{*1,2}$, Ritambhara Singh$^{*1,3}$\\
    $^{1}$Department of Computer Science, Brown University \\
    $^{2}$Institute for Bioinformatics and Medical Informatics, University of Tübingen \\
    $^{3}$Center for Computational Molecular Biology, Brown University \\
    $^{\dagger}$Co-first Authors \quad $^{*}$Co-corresponding Authors \\
    
}

\title{PiCME: Pipeline for Contrastive Modality Evaluation and Encoding in the MIMIC Dataset}
\date{}
\maketitle

\begin{abstract}
\noindent
Multimodal deep learning holds promise for improving clinical prediction by integrating diverse patient data, including text, imaging, time-series, and structured demographics. Contrastive learning facilitates this integration by producing a unified representation that can be reused across tasks, reducing the need for separate models or encoders. Although contrastive learning has seen success in vision-language domains, its use in clinical settings remains largely limited to image and text pairs. We propose the Pipeline for Contrastive Modality Evaluation and Encoding (PiCME), which systematically assesses five clinical data types from MIMIC: discharge summaries, radiology reports, chest X-rays, demographics, and time-series. We pre-train contrastive models on all 26 combinations of two to five modalities and evaluate their utility on in-hospital mortality and phenotype prediction. To address performance plateaus observed with the inclusion of more modalities, we introduce a Modality-Gated LSTM that weights each modality according to its contrastively learned importance. Our results show that contrastive models remain competitive with supervised baselines, particularly in three-modality settings. Performance declines beyond three modalities, a limitation not fully addressed by supervised models. However, the Modality-Gated LSTM mitigates this decline, improving AUROC from 73.19\% to 76.93\% and AUPRC from 51.27\% to 62.26\% in the five-modality setting. Beyond predictive performance, we compare contrastively learned modality importance scores with attribution scores and evaluate generalization across demographic subgroups, demonstrating strengths in both interpretability and fairness. PiCME is the first to scale contrastive learning across all modality combinations in MIMIC, offering insights into optimal modality selection and training strategies, and guiding future work toward more efficient and equitable clinical prediction.
\end{abstract}

\noindent
\textbf{Code:} \url{https://github.com/rsinghlab/PiCME}


\begin{figure}[h!]
    \centering
    \includegraphics[scale=0.7]{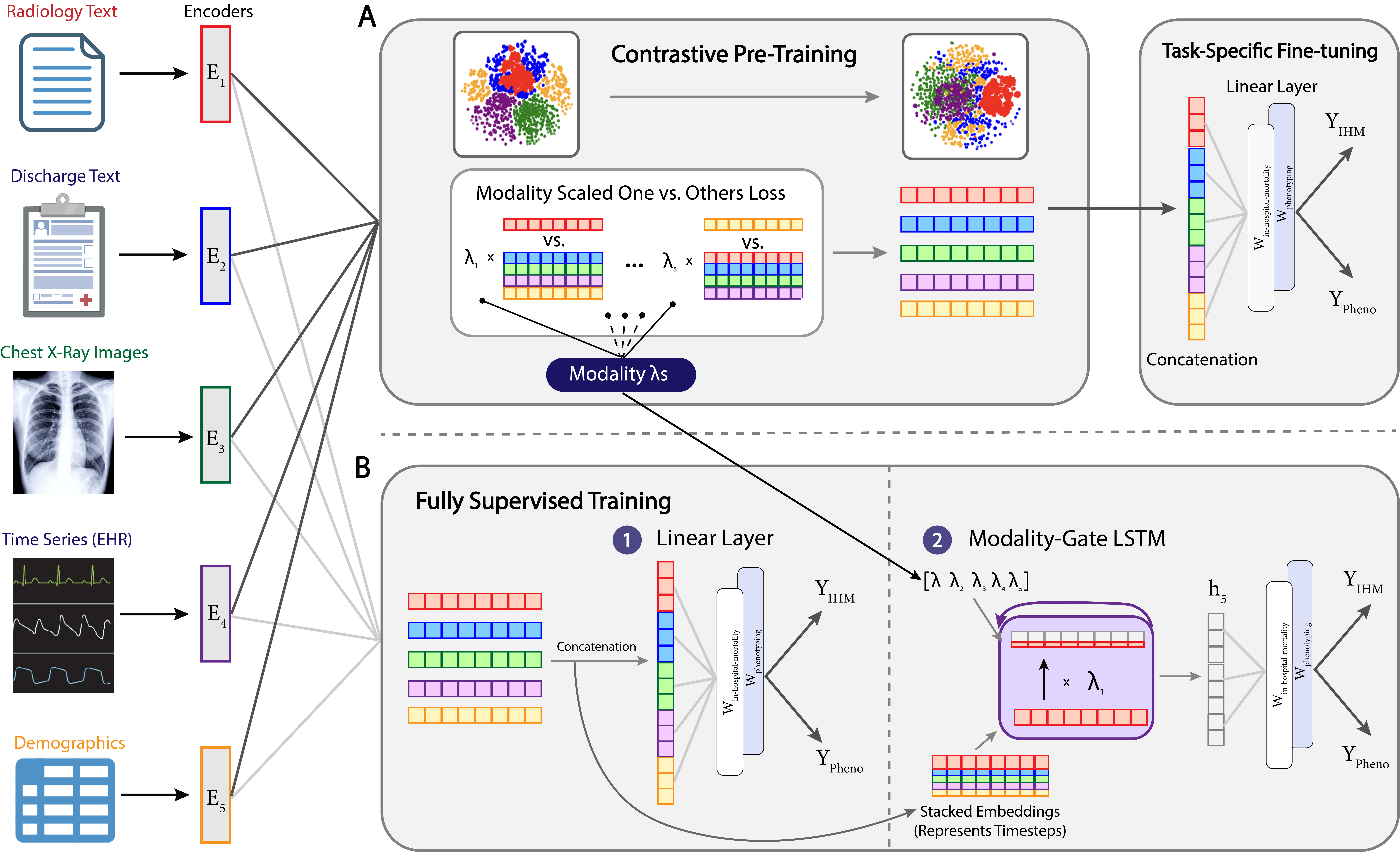}
    \caption{Overview of the MIMIC Pipeline for Contrastive Modality Evaluation (PiCME). 
    Panel (A): Contrastive learning with One-vs-Others loss, followed by fine-tuning for in-hospital mortality and phenotyping. 
    Panel (B): Fully supervised fine-tuning: (B.1) modality encoders trained per task, and (B.2) a modality gate incorporates contrastive-learned importance weights to regularize five-modality fusion.}
    \label{fig:pipeline_overview}
\end{figure}

\section{Introduction}
The availability of large-scale, de-identified clinical datasets has enabled a surge in data-driven medical research. The MIMIC dataset \cite{johnson2016mimic, johnson2023mimic} is one of the most comprehensive, with tabular demographics, time-series, patient notes, and lab measurements. The addition of MIMIC-CXR \cite{johnson2019mimic} introduced chest X-rays, enabling multimodal research that combines imaging with existing modalities from MIMIC-IV for improved clinical prediction. Contrastive learning has emerged as a favorable framework for multimodal integration due to its ability to produce reusable embeddings in a single pre-training phase, reducing reliance on task-specific supervision and improving generalizability across diverse downstream applications \cite{zhai2023sigmoid, li2021align, li2022blip}. Building on this success, contrastive learning methods have gained traction in the medical domain \cite{wang2022medclip, huang2023plip}, including in MIMIC, where they have been applied to integrate image and text modalities \cite{huang2021gloria, zhang2022contrastive, wang2022multi}.

While contrastive learning has shown promise, most studies are limited to only two modalities, typically image and text. There is no systematic evaluation of all possible modality combinations, making it unclear how different data types contribute to performance. Clinical modalities such as time-series and tabular data remain underexplored in contrastive learning, limiting the generalizability, reproducibility, and efficiency of multimodal models. Suboptimal modality choices can lead to inconsistent results and increased computational costs \cite{golovanevsky2024one}.

To address these gaps, we propose the \textbf{Pipeline for Contrastive Modality Evaluation and Encoding (PiCME)}, a framework for analyzing modality importance in contrastive learning. We systematically evaluate five clinical data types from MIMIC-IV \cite{johnson2023mimic} and MIMIC-CXR \cite{johnson2019mimic}: discharge summaries, clinical time-series, chest X-rays, radiology reports, and tabular demographics (Figure~\ref{fig:pipeline_overview}). Our goal is to identify scalable and clinically effective strategies for multimodal integration while providing insight into the relative importance of each modality. We are the first to train contrastive models on all 26 combinations of two to five modalities and compare their downstream performance on two clinical tasks: in-hospital mortality (binary classification) and phenotype prediction (25-way classification). We observe that contrastive learning reaches peak performance with three modalities: discharge notes, chest X-rays, and demographics, achieving 75.94\% AUROC on the in-hospital mortality task. The supervised baseline also peaks with the same three modalities, reaching 73.84\% AUROC in phenotype prediction. Performance declines beyond three modalities for both strategies and tasks, likely due to integration noise and feature redundancy.

To overcome the multimodal performance plateau, we introduce the \textit{Modality-Gated Long Short-Term Memory (mLSTM)}, which incorporates contrastively learned \textit{modality-$\lambda$} values to guide dynamic weighting of each modality during supervised fine-tuning (Figure~\ref{fig:pipeline_overview}A). This mechanism enables the model to selectively focus on the most informative modalities. The mLSTM surpasses the three-modality plateau, achieving 76.93\% AUROC in IHM and 74.36\% AUROC in phenotyping using all five modalities. 

Beyond predictive performance, we assess the interpretability and robustness of the learned modality weights. Using Integrated Gradients, we validate that the contrastively learned \textit{modality-$\lambda$} values align with input attributions during fine-tuning, offering insight into which modalities drive model predictions. To evaluate fairness and generalization, we examine performance across demographic subgroups including age, gender, and ethnicity. 

Overall, our work delivers a comprehensive pipeline for evaluating multimodal integration in MIMIC, offering actionable insights into modality selection, training strategy, and interpretability. This is especially important as clinical deep learning systems increasingly incorporate diverse data modalities, creating a growing need for scalable, transparent, and equitable models that generalize across patient populations. Moreover, our results demonstrate that information learned during contrastive pre-training can be effectively transferred to improve downstream supervised fine-tuning through modality-aware regularization.

\section{Related Work}

Multimodal learning has shown strong performance in both vision-language tasks \cite{li2021align, li2022blip, liu2024llavanext} and clinical decision support \cite{wang2022medclip, hayat2022medfuse, golovanevsky2022multimodal} by leveraging diverse data sources for richer representations. With the rise of large-scale unlabeled data, contrastive learning has achieved notable success, especially in vision-language models such as CLIP \cite{radford2021clip}. Building on CLIP’s foundation, contrastive learning remains central to multimodal integration, with continued advancements refining its methods \cite{zhai2023sigmoid, li2021align, li2022blip}.

Despite the success multimodal models have achieved in integrating diverse modalities, contrastive learning is rarely applied beyond the two-modality setting, primarily integrating vision and language \cite{wang2022medclip, huang2023plip, zhang2022contrastive}. This constraint arises from the pairwise design of the InfoNCE loss, which aligns positive pairs while separating negatives. To prevent a quadratically growing number of contrastive loss pairs that fail to capture inter-modality relations, a loss function that optimizes across all modalities simultaneously is needed. One successful approach is the One-Versus-Others (OvO) loss, which extends InfoNCE to handle multiple modalities in a scalable manner \cite{pototzky2022fastsiam, piran2024contrasting, thapa2024sleepfm}. Given these advantages, OvO loss is a strong candidate for contrastive training in highly multimodal datasets, such as MIMIC.

The MIMIC dataset, a key resource in medical AI research, includes MIMIC-IV (demographics, discharge notes, and EHR data) and MIMIC-CXR (chest X-rays and reports). Given the wide range of data types available, MIMIC-based multimodal studies vary in their modality choices, integration methods, and learning paradigms. \textcite{agostini2024weakly} use supervised learning for chest X-rays and notes, while \textcite{huang2021gloria}, \textcite{zhang2022contrastive}, and \textcite{wang2022multi}, apply contrastive learning to image-text pairs for both unsupervised and supervised tasks. \textcite{hayat2022medfuse} incorporates an LSTM to handle missing modalities, demonstrating its effectiveness in improving multimodal fusion. Expanding beyond two modalities, \textcite{soenksen2022integrated} integrates demographics, time series, natural language, and medical images using a fusion network, highlighting the potential of more comprehensive multimodal frameworks.

Despite the success of previous multimodal MIMIC studies, they often focus on limited and arbitrary modality selection and lack systematic evaluations of contrastive learning beyond two modalities. In MIMIC, different modality combinations can yield varying results, yet predictive tasks rarely offer modality-level interpretability. Furthermore, modality importance may differ between task-agnostic pre-training and task-specific fine-tuning, yet this distinction is rarely explored. This is a critical knowledge gap because without understanding how each modality contributes across stages, model design becomes inconsistent and uninformed, limiting generalizability, interpretability, and efficient use of clinical data. We address these gaps by evaluating all possible modality configurations and analyzing modality contributions across both pre-training and fine-tuning.


\section{Methods}
\subsection{Modality Encoders}\label{sec:modality_encoders}

We integrate five modalities to enhance clinical classification: discharge summaries, radiology reports, chest X-rays, clinical time-series, and demographic data. To ensure comparability with prior research, we adopt encoder architectures that have shown strong performance in previous MIMIC frameworks \cite{hayat2022medfuse, soenksen2022integrated, huang2021gloria, wang2022multi}.

Discharge summaries capture detailed hospitalization events, while radiology reports contain expert image interpretations. We encode both using ClinicalBERT \cite{alsentzer2019publicly}, a BERT \cite{devlin2018bert} variant pre-trained on clinical text and widely used in multimodal MIMIC studies \cite{soenksen2022integrated, liu2023attention}. To improve efficiency, we fine-tune ClinicalBERT using LoRA, a low-rank adaptation method that limits the number of trainable parameters \cite{hu2022lora}.

Chest X-ray images are encoded with a ResNet-based model \cite{he2016deep}, commonly used in MIMIC-CXR for its strong feature extraction \cite{hayat2022medfuse}. Clinical time-series data, consisting of ICU-recorded vitals and lab results, is modeled using an LSTM \cite{hochreiter1997long}, which captures temporal patterns in sequential medical data \cite{hayat2022medfuse, soenksen2022integrated}. Demographics data, which includes age, gender, and ethnicity, is encoded using a multi-layer perceptron (MLP), a standard approach for structured categorical and numerical features \cite{soenksen2022integrated}.

Each encoder transforms raw patient data into a modality-specific feature representation, used in both contrastive pre-training, supervised fine-tuning, and fully-supervised baselines. During contrastive pre-training, encoders align multimodal features via self-supervised objectives and are later fine-tuned for specific tasks. For fully-supervised baselines, all encoders are trained from scratch to establish a non-contrastive baseline.

\subsection{Unsupervised Contrastive Pre-Training}

To learn robust multimodal representations, we use contrastive learning, a self-supervised method that encourages similar data points to be closer in the latent space while pushing dissimilar ones apart. We explore two contrastive learning approaches: pairwise contrastive learning using the InfoNCE loss and multimodal contrastive learning using One-Versus-Others (OvO) loss \cite{pototzky2022fastsiam}. For cases involving exactly two modalities, we use the standard pairwise InfoNCE loss, whereas for three or more modalities, we adopt OvO loss, a non-pairwise method that creates one loss term per modality. In the two-modality case, OvO loss reduces to InfoNCE loss since there are only two modalities, leaving no additional modalities to average over when forming the "others" representation. Additionally, we introduce a learnable weighting scheme in the OvO loss to account for modality importance dynamically.

\subsubsection{Pairwise Contrastive Learning with InfoNCE}
The InfoNCE loss \cite{oord2018representation} is widely used in contrastive learning frameworks such as SimCLR \cite{chen2020simple}, MoCo \cite{he2020momentum}, and CLIP \cite{radford2021learning}. It is designed to pull together positive pairs \emdash different views of the same data point \emdash while pushing apart negative pairs, which are other samples in the batch. 

For each sample $k$ in a batch, let $x_k^i$ and $x_k^j$ represent the embeddings from two different modalities, $i$ and $j$, respectively. The pairwise contrastive prediction loss is defined as:

\begin{equation}
    \mathcal{L}_{\text{pair}}^{i,j,k} = -\log \frac{\exp (\operatorname{sim}(x_k^i, x_k^j) / \tau )}{\sum_{m=1}^{N} \exp (\operatorname{sim}(x_k^i, x_m^j) / \tau )}
\end{equation}

where $N$ is the number of samples in a batch, $\tau$ is a trainable temperature parameter, and $\operatorname{sim}(\cdot, \cdot)$ represents cosine similarity. The final InfoNCE loss for a batch sums over all pairs of modalities $(i, j)$ across all samples:

\begin{equation}
    \mathcal{L}_{\text{InfoNCE}} = \sum_{i,j} \sum_{k=1}^{N} \mathcal{L}_{\text{pair}}^{i,j,k}.
\end{equation}

This formulation ensures that embeddings from the same data sample but across different modalities are drawn closer, while embeddings from different samples remain separated. However, as the number of modalities increases, each loss term is only representative of the relationship between two modalities, not learning how one modality relates to multiple modalities at once. To address this, we adopt a One-Versus-Others loss, which extends contrastive learning to settings with more modalities.

\subsubsection{multimodal Contrastive Learning with One-Versus-Others (OvO) Loss}

For cases involving three or more modalities, we move beyond the pairwise paradigm and instead use the One-Versus-Others (OvO) loss, also known as One-Versus-Rest or Leave-One-Out Loss~\cite{pototzky2022fastsiam, piran2024contrasting, thapa2024sleepfm}. This method has been successfully applied in attention mechanisms~\cite{golovanevsky2024one} and has shown strong performance in learning multi-view representations in contrastive learning.

In contrast to InfoNCE, which enforces similarity between pairs of modalities, OvO encourages each modality to align with the collective representation of all other modalities. For each sample $k$ in the batch and each modality $i$, we define the ``other'' embedding by averaging over embeddings from all remaining modalities:

\begin{equation}
    \bar{x}_{k}^{\neg i} = \frac{1}{K - 1} \sum_{\substack{j=1 \\ j \neq i}}^{K} x_k^j.
\end{equation}

The OvO contrastive loss is then defined as:

\begin{equation}
    \mathcal{L}_{\text{OvO}}^{i,k} = -\log \frac{\exp (\operatorname{sim}(x_k^i, \bar{x}_k^{\neg i}) / \tau )}{\sum_{m=1}^{N} \exp (\operatorname{sim}(x_k^i, \bar{x}_m^{\neg i}) / \tau )}.
\end{equation}

To allow the model to learn the relative importance of each modality dynamically, we extend the OvO loss by introducing modality-specific learnable mixture weights $\lambda_i$ for each modality $i$. Instead of treating all modalities equally, the model learns to assign higher importance to more informative modalities while down-weighting less relevant ones. The weighted OvO loss is defined as:

\begin{equation}
    \mathcal{L}_{\text{weighted-OvO}}^{i,k} = -\lambda_i \log \frac{\exp (\operatorname{sim}(x_k^i, \bar{x}_k^{\neg i}) / \tau )}{\sum_{m=1}^{N} \exp (\operatorname{sim}(x_k^i, \bar{x}_m^{\neg i}) / \tau )},
\end{equation}

where each $\lambda_i$ is a trainable parameter, optimized jointly with the contrastive objective. After each step, the $\lambda$ values are normalized using the softmax function such that $\sum_i \lambda_i = 1$. The final weighted loss is:

\begin{equation}
    \mathcal{L}_{\text{weighted-OvO}} = \sum_{i=1}^{K} \sum_{k=1}^{N} \mathcal{L}_{\text{weighted-OvO}}^{i,k}.
\end{equation}

This modification allows the model to automatically adjust modality contributions based on their usefulness in the learning task. These weights act as a regularization mechanism, adjusting the contribution of each modality to enhance model robustness.

By combining standard contrastive learning for two modalities with a modality-weighted OvO-based contrastive approach for three or more modalities, our method effectively captures both pairwise interactions and global multimodal dependencies, ensuring robust representation learning across diverse clinical modalities.

\subsection{Supervised Fine-Tuning}

\subsubsection{Predictive Tasks and Training Objectives}
We evaluate our approach on two clinical tasks: \textbf{in-hospital mortality} prediction (IHM) and \textbf{phenotyping}. The IHM prediction task is a binary classification problem that aims to predict whether a patient will survive their hospital stay based on data from the first 48 hours of an ICU admission (following \cite{hayat2022medfuse}).

The phenotype classification task is a multi-label classification problem, where the goal is to predict whether a patient has any of 25 chronic, mixed, or acute care conditions during an ICU stay. For both tasks, we evaluate performance using the Area Under the Receiver Operating Characteristic (AUROC) curve and the Area Under the Precision-Recall Curve (AUPRC), following \textcite{liu2023attention, hayat2022medfuse, soenksen2022integrated}. The full details on hyperparameter tuning and data splits is in Appendix Section \ref{app:exp_set}.

\subsubsection{Contrastive Pre-Training for Downstream Tasks}
For contrastive learning, each modality is first passed through a dedicated encoder to produce embeddings, which are optimized using a contrastive loss. After pre-training, the resulting embeddings are used for downstream classification tasks, including in-hospital mortality and phenotype prediction (see Figure~\ref{fig:pipeline_overview}A).

To integrate multiple modalities, we concatenate the contrastively trained embeddings to form a unified representation, similar to common fusion strategies used in prior work such as HAIM~\cite{soenksen2022integrated} and MedFuse~\cite{hayat2022medfuse}. The final representation is defined as:
\begin{equation}
x \in \mathbb{R}^{nm}
\end{equation}
where \( n \) is the embedding size and \( m \) is the number of modalities. This fused vector is passed to a multi-layer perceptron (MLP) for task-specific prediction.

\subsubsection{Fully Supervised Baseline}
As a comparison, we implement fully supervised baselines for each modality combination. These models omit contrastive pre-training and train all modality encoders from scratch. The architecture mirrors the contrastive setup, using the same modality encoders (Section~\ref{sec:modality_encoders}) and a shared classification head that takes concatenated embeddings as input (see Figure~\ref{fig:pipeline_overview} B.1).

This setup allows us to isolate the benefits of contrastive pre-training by comparing models trained end-to-end with those using fixed, pretrained embeddings.

\subsubsection{Modality-Gated LSTM}
We introduce the Modality-Gated LSTM (mLSTM), an extension of the Long Short-Term Memory (LSTM) neural network. LSTMs are a type of recurrent neural network designed to capture long-range dependencies in sequential data by using gated mechanisms to regulate information flow. In the mLSTM, the \textit{cell state}—a key component of LSTMs that retains long-term dependencies—is dynamically adjusted using modality-specific importance weights (\(\lambda_i\)) learned during contrastive pre-training. This allows the network to modulate the contribution of each modality when making predictions, ensuring that the influence of each input aligns with its learned relevance from the contrastive training step. 

The mLSTM follows the standard formulation of the LSTM featuring the input gate ($I_t$), forget gate ($F_t$), candidate memory ($\widetilde{C}_t$), and output gate ($O_t$). To introduce modality-enhanced updates, we incorporate a modality-specific scaling term $\lambda_i$ for each modality $i$ where $\sum_i^{|\Lambda|}\lambda_i = 1$. Here, the $\lambda$ parameters are learned during contrastive pre-training, but can be substituted with predefined modality importance values. This value adjusts the contribution of the candidate memory, $C_t$ with respect to $\lambda_t$, to the calculation of the cell ($C_t$) and hidden states ($H_t$) for timestep $t$:

\begin{equation*}
\begin{aligned}
    \vec{\lambda}_t &= \langle \lambda_t, ..., \lambda_t\rangle, |\vec{\lambda}_t| = h, \\
    C_t &= F_t \odot C_{t-1} + (I_t \odot \widetilde{C}_t) \odot  \vec{\lambda}_t \\
    H_t &= O_t \odot \text{tanh}(C_t)
\end{aligned}
\end{equation*}

The final hidden state $H_t \in \mathbb{R}^h$ from the model is then fed into a standard MLP for task-specific predictions. This fusion and classification method is only used for fully supervised models.

\section{Results}
We present our results in two parts. First, we examine contrastive learning as a task-agnostic framework by evaluating how well different combinations of clinical modalities align in a shared representation space. We analyze the impact of scaling to more modalities and observe diminishing returns in alignment quality. Second, we focus on task-specific prediction for in-hospital mortality and phenotype classification. We observe that both contrastive and supervised baselines reach peak performance with three modalities and begin to plateau or decline as more are added. To overcome this performance plateau, we introduce the Modality-Gated LSTM, which uses contrastively learned modality importance to improve supervised fine-tuning.

\subsection{More Modalities, Less Alignment: Challenges in Task-Agnostic Contrastive Learning}\label{sec:contrastive_results}
\begin{wrapfigure}{r}{0.5\textwidth}
    \centering
    \vspace{-10pt}  
    \includegraphics[width=0.397\textwidth]{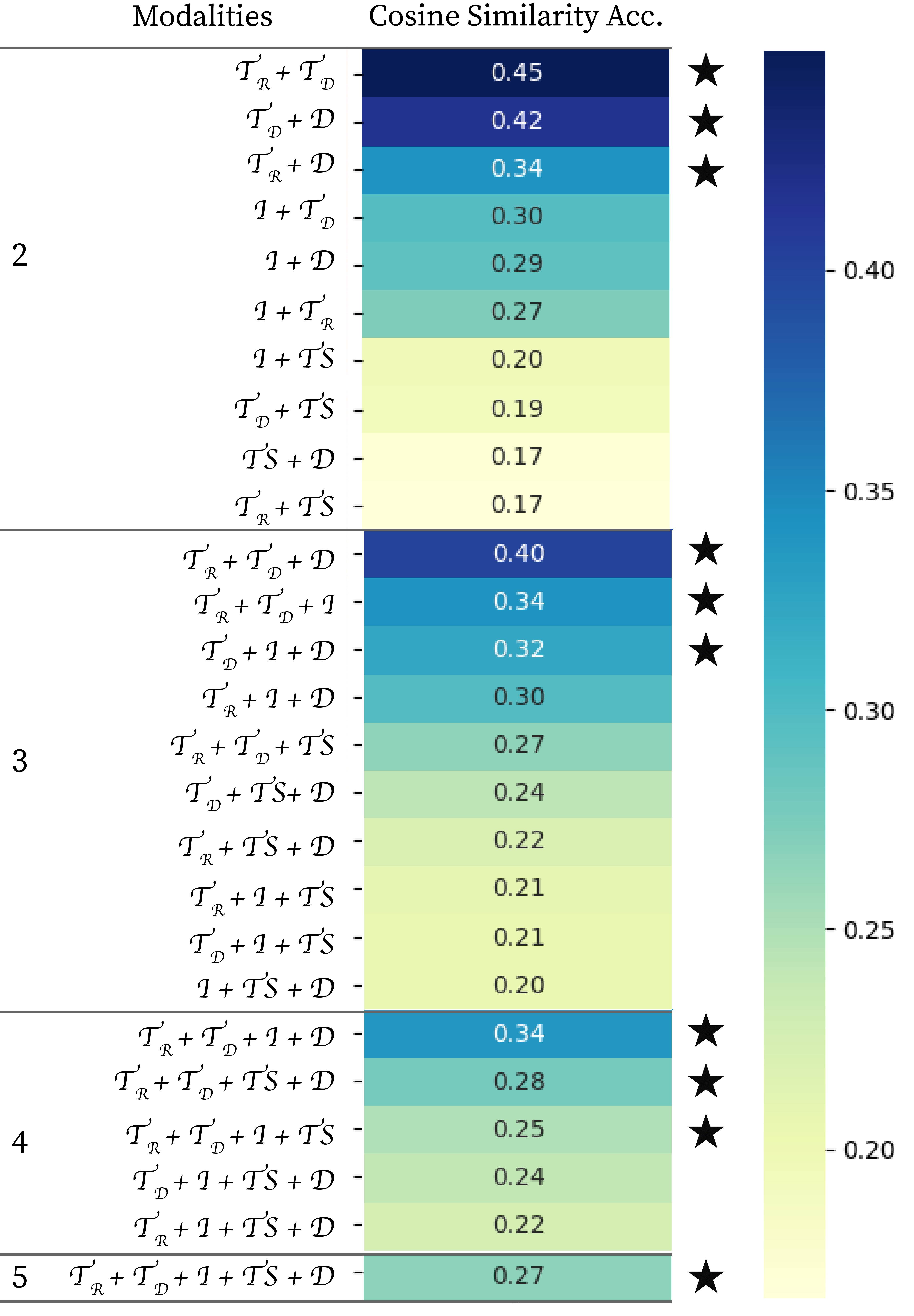}
    \caption{Top-5 cosine similarity accuracy for each modality combination. We use five modalities: Radiology Text (\(\mathcal{T}_{\mathcal{R}}\)), Discharge Text (\(\mathcal{T}_{\mathcal{D}}\)), Images (\(\mathcal{I}\)), Demographics (\(\mathcal{D}\)), and Time-Series (\(\mathcal{TS}\)).}
    \label{fig:CL_accuracies}
    \vspace{-10pt}  
\end{wrapfigure}
\begin{figure}
    \centering\includegraphics[scale=0.3]{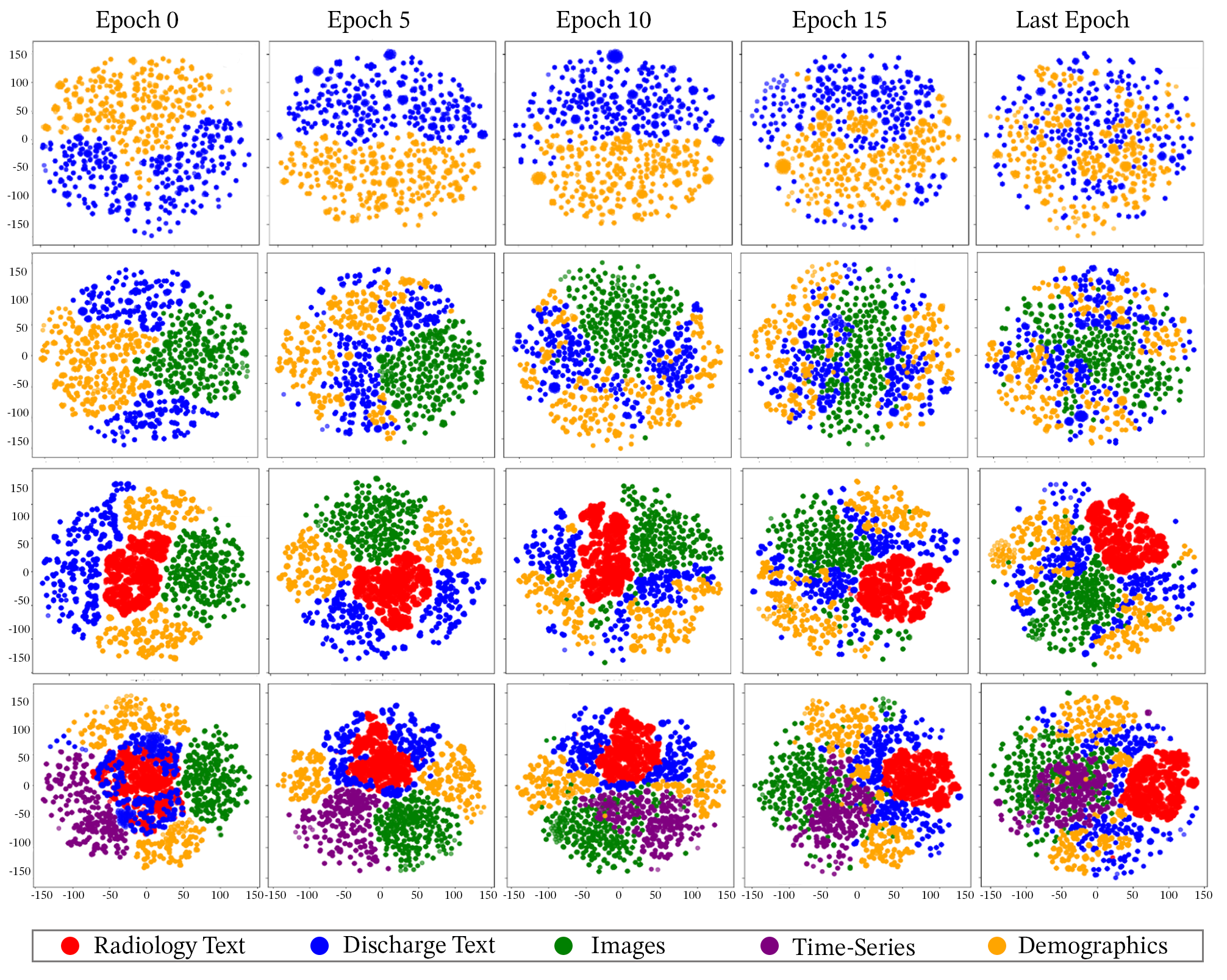}
    \caption{T-SNE visualization of contrastive learning through epochs.}\label{fig:CL_tsne}
\end{figure}

To visualize the effect of contrastive pre-training, we use t-SNE to project high-dimensional modality embeddings into two dimensions. t-SNE is well-suited for capturing local structure, making it effective for observing how modality embeddings align during training, unlike PCA or UMAP which emphasize global variance or manifold preservation. Figure~\ref{fig:CL_tsne} shows how embeddings from two to five modalities gradually merge over successive training epochs. Each point represents an embedding from the modality encoders in Section~\ref{sec:modality_encoders}. 
The merging of modalities is most pronounced in the two- and three-modality settings. In contrast, four- and five-modality combinations show weaker alignment, with modality clusters remaining more dispersed. Radiology notes, in particular, remain distinct throughout training, suggesting that their domain-specific language makes them harder to align with structured or temporal data such as demographics or time-series.
To complement this qualitative view, we compute a quantitative alignment score using top-5 accuracy based on cosine similarity in the shared embedding space. A sample is considered correctly aligned if at least one of its five nearest neighbors comes from the same patient, regardless of modality. This metric reflects how consistently different modalities from the same patient are mapped close together.

Figure~\ref{fig:CL_accuracies} shows a clear trend: as the number of modalities increases, top-5 similarity declines. The highest accuracy, at 45\%, is observed with radiology and discharge text. Among three-modality combinations, the top-performing set reaches 40\% accuracy with radiology text, discharge notes, and demographics. We hypothesize that this is due to both text modalities being encoded using ClinicalBERT, which provides a shared latent space that facilitates alignment. Beyond text-to-text alignment, demographics is a highly adaptable modality because its low-dimensional, structured representation (e.g., sex, age, ethnicity, marital status) can be easily mapped into the shared embedding space, which may explain its higher similarity scores when combined with other modalities.

By contrast, time-series data consistently reduces alignment quality. Their high variance and temporal dynamics make them more difficult to reconcile with static or text-based modalities. When included in two- or three-modality settings, time-series embeddings lower cosine similarity scores, indicating challenges in forming coherent joint representations. 

These findings underscore the difficulty of scaling contrastive learning to high-dimensional multimodal spaces, particularly when attempting to align more than three modalities. The modality combinations with the strongest alignment in Figure \ref{fig:CL_accuracies} are used in downstream fine-tuning, as they offer the most promising foundations for task-specific learning.


\subsection{Effectiveness of Contrastive Pre-training and Modality-Gated LSTM on Task-Specific Fine-Tuning}
\begin{table}[ht]
\centering
\begin{tabular}{>{\centering\arraybackslash}m{3.8cm}c|cc|cc} 
\toprule
\multirow{2}{*}{\textbf{Modalities}} & \multirow{2}{*}{\textbf{Approach}} & \multicolumn{2}{c|}{\textbf{In-Hospital Mortality}} & \multicolumn{2}{c}{\textbf{Phenotyping}} \\
 & & \textbf{AUROC} & \textbf{AUPRC} & \textbf{AUROC} & \textbf{AUPRC} \\
\midrule
$\mathcal{T}_{\mathcal{R}}$ + $\mathcal{T}_{\mathcal{D}}$ & Contrastive & 57.52 {\tiny$\pm$ 1.71} & 27.25 {\tiny$\pm$ 1.75} & 68.90 {\tiny$\pm$ 0.08} & 43.25 {\tiny$\pm$ 0.45} \\
$\mathcal{T}_{\mathcal{R}}$ + $\mathcal{T}_{\mathcal{D}}$ & Supervised Baseline & 58.70 {\tiny$\pm$ 2.30} & 31.30 {\tiny$\pm$ 2.98} & 65.09 {\tiny$\pm$ 0.77} & 38.05 {\tiny$\pm$ 0.78} \\
\textbf{$\mathcal{T}_{\mathcal{D}}$ + $\mathcal{D}$} & \textbf{Contrastive} & \textbf{64.29} {\tiny$\pm$ 0.91} & \textbf{37.56} {\tiny$\pm$ 0.58} & 62.81 {\tiny$\pm$ 0.17} & 36.08 {\tiny$\pm$ 0.24} \\
$\mathcal{T}_{\mathcal{D}}$ + $\mathcal{D}$ & \textbf{Supervised Baseline} & 61.02 {\tiny$\pm$ 2.92} & 28.80 {\tiny$\pm$ 2.09} & \textbf{73.76} {\tiny$\pm$ 0.88} & \textbf{48.20} {\tiny$\pm$ 1.06} \\
$\mathcal{T}_{\mathcal{R}}$ + $\mathcal{D}$ & Contrastive & 61.34 {\tiny$\pm$ 0.25} & 27.93 {\tiny$\pm$ 0.48} & 64.10 {\tiny$\pm$ 0.09} & 36.20 {\tiny$\pm$ 0.25} \\
$\mathcal{T}_{\mathcal{R}}$ + $\mathcal{D}$ & Supervised Baseline & 63.50 {\tiny$\pm$ 1.58} & 30.50 {\tiny$\pm$ 1.25} & 72.87 {\tiny$\pm$ 0.94} & 45.47 {\tiny$\pm$ 0.98} \\
\midrule
$\mathcal{T}_{\mathcal{R}}$ + $\mathcal{T}_{\mathcal{D}}$ + $\mathcal{D}$ & Contrastive & 61.23 {\tiny$\pm$ 2.43} & 28.20 {\tiny$\pm$ 1.81} & 68.32 {\tiny$\pm$ 0.11} & 42.61 {\tiny$\pm$ 0.22} \\
$\mathcal{T}_{\mathcal{R}}$ + $\mathcal{T}_{\mathcal{D}}$ + $\mathcal{D}$ & Supervised Baseline & 62.82 {\tiny$\pm$ 2.80} & 29.65 {\tiny$\pm$ 2.15} & 73.40 {\tiny$\pm$ 1.00} & 46.13 {\tiny$\pm$ 1.30} \\
$\mathcal{T}_{\mathcal{R}}$ + $\mathcal{T}_{\mathcal{D}}$ + $\mathcal{I}$ & Contrastive & 63.97 {\tiny$\pm$ 5.50} & 32.88 {\tiny$\pm$ 5.20} & 62.24 {\tiny$\pm$ 0.50} & 35.11 {\tiny$\pm$ 0.53} \\
$\mathcal{T}_{\mathcal{R}}$ + $\mathcal{T}_{\mathcal{D}}$ + $\mathcal{I}$ & Supervised Baseline & 66.77 {\tiny$\pm$ 3.05} & 35.59 {\tiny$\pm$ 3.33} & 72.36 {\tiny$\pm$ 0.50} & 46.38 {\tiny$\pm$ 0.41} \\
$\mathcal{T}_{\mathcal{D}}$ + $\mathcal{I}$ + $\mathcal{D}$ & \textbf{Contrastive} & \textbf{75.94} {\tiny$\pm$ 0.50} & \textbf{58.32} {\tiny$\pm$ 2.16} & 62.93 {\tiny$\pm$ 0.14} & 36.15 {\tiny$\pm$ 0.29} \\
$\mathcal{T}_{\mathcal{D}}$ + $\mathcal{I}$ + $\mathcal{D}$ & \textbf{Supervised Baseline} & 65.50 {\tiny$\pm$ 1.94} & 39.14 {\tiny$\pm$ 3.93} & \textbf{73.84} {\tiny$\pm$ 0.26} & \textbf{49.34} {\tiny$\pm$ 0.57} \\
\midrule
$\mathcal{T}_{\mathcal{R}}$ + $\mathcal{T}_{\mathcal{D}}$ + $\mathcal{I}$ + $\mathcal{D}$ & Contrastive & 62.98 {\tiny$\pm$ 2.52} & 29.69 {\tiny$\pm$ 2.13} & 67.21 {\tiny$\pm$ 0.23} & 40.08 {\tiny$\pm$ 0.40} \\
$\mathcal{T}_{\mathcal{R}}$ + $\mathcal{T}_{\mathcal{D}}$ + $\mathcal{I}$ + $\mathcal{D}$ & Supervised Baseline & 66.92 {\tiny$\pm$ 3.04} & 38.24 {\tiny$\pm$ 5.67} & 73.13 {\tiny$\pm$ 1.19} & 48.17 {\tiny$\pm$ 0.42} \\
\textbf{$\mathcal{T}_{\mathcal{R}}$ + $\mathcal{T}_{\mathcal{D}}$ + $\mathcal{I}$ + $\mathcal{D}$} & \textbf{Modality-Gated LSTM} & 62.82 {\tiny$\pm$ 7.47} & 32.81 {\tiny$\pm$ 4.36} & \textbf{73.32} {\tiny$\pm$ 0.52} & 50.51 {\tiny$\pm$ 1.16} \\
$\mathcal{T}_{\mathcal{R}}$ + $\mathcal{T}_{\mathcal{D}}$ + $\mathcal{TS}$ + $\mathcal{D}$ & Contrastive & 64.45 {\tiny$\pm$ 0.52} & 34.36 {\tiny$\pm$ 0.45} & 68.69 {\tiny$\pm$ 0.10} & 42.76 {\tiny$\pm$ 0.24} \\
$\mathcal{T}_{\mathcal{R}}$ + $\mathcal{T}_{\mathcal{D}}$ + $\mathcal{TS}$ + $\mathcal{D}$ & Supervised Baseline & 73.02 {\tiny$\pm$ 2.35} & 50.49 {\tiny$\pm$ 3.82} & 72.45 {\tiny$\pm$ 0.98} & 45.32 {\tiny$\pm$ 0.38} \\
\textbf{$\mathcal{T}_{\mathcal{R}}$ + $\mathcal{T}_{\mathcal{D}}$ + $\mathcal{TS}$ + $\mathcal{D}$} & Modality-Gated LSTM & 73.70 {\tiny$\pm$ 3.13} & 50.22 {\tiny$\pm$ 2.58} & 72.81 {\tiny$\pm$ 0.34} & 49.72 {\tiny$\pm$ 0.48} \\
$\mathcal{T}_{\mathcal{R}}$ + $\mathcal{T}_{\mathcal{D}}$ + $\mathcal{I}$ + $\mathcal{TS}$ & Contrastive & 63.18 {\tiny$\pm$ 0.94} & 31.70 {\tiny$\pm$ 0.63} & 68.50 {\tiny$\pm$ 0.27} & 41.92 {\tiny$\pm$ 0.44} \\
$\mathcal{T}_{\mathcal{R}}$ + $\mathcal{T}_{\mathcal{D}}$ + $\mathcal{I}$ + $\mathcal{TS}$ & Supervised Baseline & 71.02 {\tiny$\pm$ 3.62} & 39.41 {\tiny$\pm$ 3.97} & 72.91 {\tiny$\pm$ 0.44} & 46.98 {\tiny$\pm$ 0.44} \\
$\mathcal{T}_{\mathcal{R}}$ + $\mathcal{T}_{\mathcal{D}}$ + $\mathcal{I}$ + $\mathcal{TS}$ & \textbf{Modality-Gated LSTM} & \textbf{73.73} {\tiny$\pm$ 2.38} & \textbf{54.32} {\tiny$\pm$ 3.96} & 73.21{\tiny$\pm$ 0.61} & \textbf{50.52} {\tiny$\pm$ 0.11} \\
\midrule
$\mathcal{T}_{\mathcal{R}}$ + $\mathcal{T}_{\mathcal{D}}$ + $\mathcal{I}$ + $\mathcal{TS}$ + $\mathcal{D}$ & Contrastive & 61.44 {\tiny$\pm$ 3.42} & 30.88 {\tiny$\pm$ 2.37} & 68.38 {\tiny$\pm$ 0.28} & 42.65 {\tiny$\pm$ 0.46} \\
$\mathcal{T}_{\mathcal{R}}$ + $\mathcal{T}_{\mathcal{D}}$ + $\mathcal{I}$ + $\mathcal{TS}$ + $\mathcal{D}$ & Supervised Baseline & 73.19 {\tiny$\pm$ 2.62} & 51.27 {\tiny$\pm$ 6.65} & 73.43 {\tiny$\pm$ 0.52} & 47.44 {\tiny$\pm$ 0.51} \\
\rowcolor{yellow!30}
\textbf{$\mathcal{T}_{\mathcal{R}}$ + $\mathcal{T}_{\mathcal{D}}$ + $\mathcal{I}$ + $\mathcal{TS}$ + $\mathcal{D}$} & \textbf{Modality-Gated LSTM} & \textbf{76.93} {\tiny$\pm$ 2.82} & \textbf{62.26} {\tiny$\pm$ 2.79} & \textbf{74.36} {\tiny$\pm$ 0.71} & \textbf{51.21} {\tiny$\pm$ 0.77} \\
\hline
\end{tabular}
\caption{Fine-tuning Results for In-Hospital Mortality and Phenotyping Prediction. 
We use five modalities: Radiology Text (\(\mathcal{T}_{\mathcal{R}}\)), Discharge Text (\(\mathcal{T}_{\mathcal{D}}\)), Images (\(\mathcal{I}\)), Demographics (\(\mathcal{D}\)), and Time-Series (\(\mathcal{TS}\)). \label{tab:combined_results}
The final values are computed using the average of 10 random initializations to ensure robustness.}
\label{tab:results}
\end{table}

In this section, we evaluate the performance of fine-tuned models across modality combinations for In-Hospital Mortality (IHM) and phenotype prediction. Table~\ref{tab:combined_results} reports AUROC and AUPRC scores for three settings: contrastive pre-training with frozen encoders, fully-supervised fine-tuning from scratch (supervised baseline), and the proposed Modality-Gated LSTM, which uses contrastively learned modality weights during fully-supervised training.

We find that contrastively pre-trained embeddings, obtained from encoders that are frozen during fine-tuning, often match or outperform fully-supervised models when applied to fewer modalities (three and under). 

The strongest contrastive performance appears in the three-modality combination of discharge text, images, and demographics ($\mathcal{T}_{\mathcal{D}}$ + $\mathcal{I}$ + $\mathcal{D}$), achieving 75.94\% AUROC and 58.32\% AUPRC for IHM, higher than any two or three-modality fully-finetuned counterpart. This demonstrates the strength of contrastive learning when a few semantically aligned modalities are integrated. The advantage is particularly pronounced in IHM, a binary classification task, while phenotyping, as a 25-class task, appears more challenging for contrastive representations. 

As more modalities are added, contrastive performance begins to degrade. In IHM, AUROC steadily decreases from 75.94\% with three modalities to 61.44\% with five. A similar trend is observed for phenotyping AUROC scores from 68.90\% to 68.38\% along with a plateau in AUPRC. These diminishing returns likely result from redundancy and noise introduced by additional modalities, which contrastive pre-training alone does not resolve. This trend aligns with Section~\ref{sec:contrastive_results} and Figure~\ref{fig:CL_tsne}, where alignment across modalities became more difficult as the number of modalities increases.

Supervised baseline models perform slightly better than contrastive ones in four and five-modality settings, but they also show signs of saturation. For instance, the best supervised baseline IHM AUROC in a four-modality setup is 73.02\%, and in five modalities, 73.19\%, both falling short of the three-modality contrastive result of 75.94\%. 

To overcome these limitations in the four and five-modality setting, we propose the Modality-Gated LSTM, which uses contrastively learned modality importances (Table~\ref{tab:ig_scores}) to regulate the contribution of each modality during integration. For IHM, it reaches 76.93\% AUROC and 62.26\% AUPRC; for phenotyping, 74.36\% AUROC and 51.21\% AUPRC. These scores represent the highest across all models and combinations, highlighting the importance of structured integration when working with complex multimodal inputs.

Taken together, these findings offer a valuable reference for the research community. Depending on the task and modality count, Table~\ref{tab:combined_results} can guide decisions on whether contrastive pre-training or full supervision is more appropriate, and when advanced integration strategies are necessary. For binary tasks and two to three modalities, contrastive pre-training is often sufficient and, in some cases, even outperforms fully supervised models while requiring only a single shared embedding model that can be reused across tasks. This reduces the need for repeated training and large-scale labeled data. For more complex, multi-class tasks and higher modality counts, the Modality-Gated LSTM mitigates redundancy and extracts useful signal, leveraging contrastively learned modality importance to guide supervised training. This pipeline has practical implications for multimodal system design, particularly in clinical applications like MIMIC, where balancing data richness with model complexity and training efficiency is essential.

\subsection{Analyzing Model Performance Across Demographics and Modalities}
\begin{figure}[ht]
    \centering
    \includegraphics[width=\textwidth]{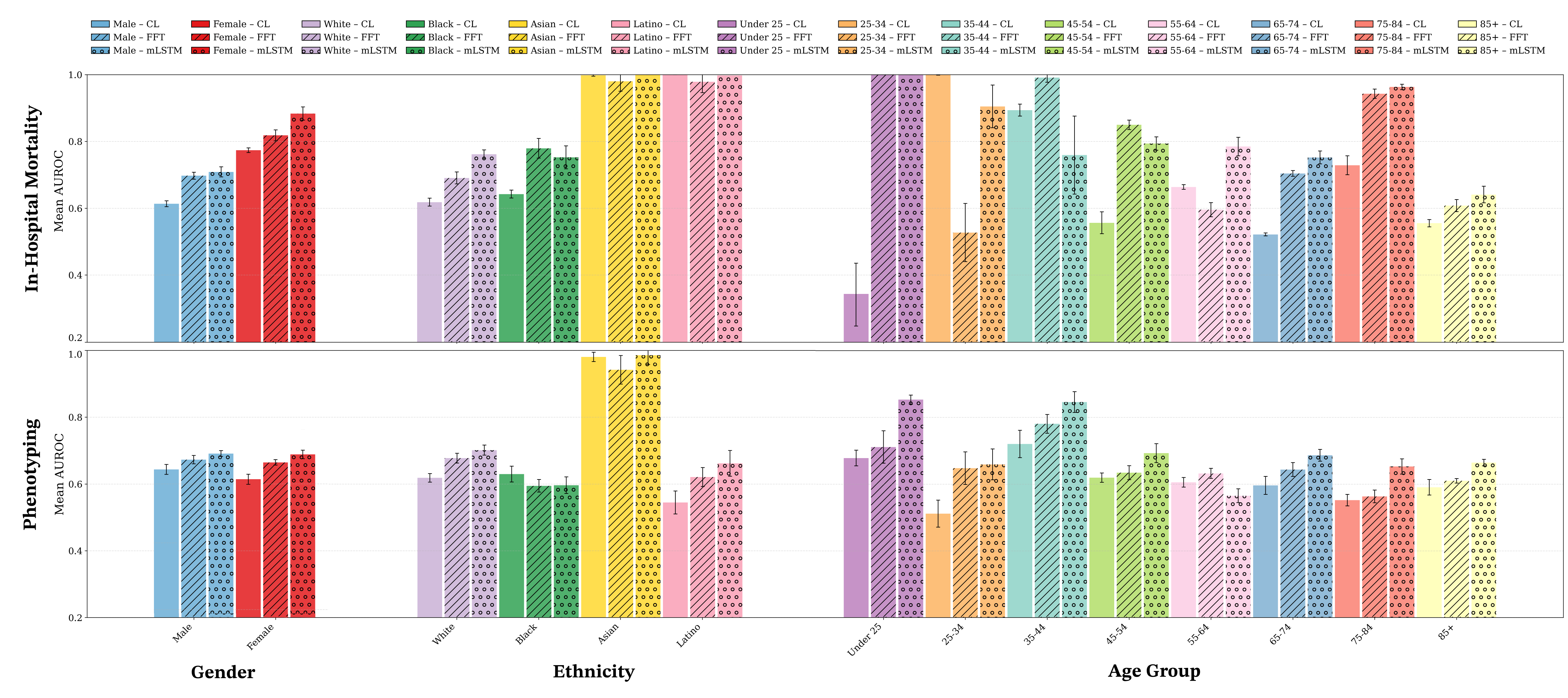}
    \caption{AUROC scores of the five-modality models across gender, ethnicity, and age group for In-Hospital Mortality (top) and Phenotyping (bottom). Models compared include Contrastive Learning Pretraining (CL), Fully Fine-Tuned (FFT), and Modality-Gated LSTM (mLSTM).}
    \label{fig:demo_performance}
\end{figure}

\subsubsection{Demographic Performance Across Models}

Building on overall trends, we analyze model performance across age, gender, and ethnicity to assess whether strong average results generalize across subpopulations. This reveals consistency, subgroup-specific strengths, and broader fairness across modeling strategies. Figure~\ref{fig:demo_performance} shows AUROC of five-modality models across demographic groups for IHM and phenotyping. We compare contrastive learning pretraining (CL), fully fine-tuned baseline (FFT), and the Modality-Gated LSTM (mLSTM). Subgroup sizes are detailed in Appendix Table~\ref{tab:demo_distribution}.

Across nearly all subgroups and tasks, mLSTM achieves the highest AUROC, suggesting that dynamic modality weighting improves accuracy and generalization. Models perform better on females in IHM despite lower representation (866 vs. 1164 males), a disparity less evident in phenotyping. mLSTM leads in both tasks and genders, showing its potential to reduce gender-specific performance gaps.

Ethnicity reveals nuanced patterns. CL shows gains for underrepresented groups in phenotyping, particularly Asian and Black patients. However, in IHM, mLSTM performs more consistently in smaller subgroups, such as Asian and Latino patients ($n=51$). Patients under 25 show lower and more variable performance, likely due to small sample size and life-stage variability. CL performs well in middle-aged groups (e.g., 55–64), while mLSTM remains stable across age brackets.

In summary, mLSTM is the most robust method across subgroups, combining high performance with consistent generalization. While CL shows promise for some populations, its instability in underrepresented or variable groups highlights the need for adaptive integration in fully supervised models.

\subsubsection{Validating Modality Lambdas with Integrated Gradients}

\begin{table}[h!]
    \centering
    \caption{Integrated Gradient (IG) Scores for each modality across the two predictive tasks.}
    \label{tab:ig_scores}
    \begin{tabular}{lccc}
        \toprule
        \textbf{Modality} & \textbf{\textit{modality-$\lambda$}}& \textbf{In-Hospital Mortality} & \textbf{Phenotyping} \\
        \midrule
        Discharge Text (\(\mathcal{T}_{\mathcal{D}}\)) & \ 0.297 &\(\text{IG}_{\mathcal{T}_{\mathcal{D}}} = 0.264\) & \(\text{IG}_{\mathcal{T}_{\mathcal{D}}} = 0.336\) \\
        Radiology Text (\(\mathcal{T}_{\mathcal{R}}\)) & \ 0.245 &\(\text{IG}_{\mathcal{T}_{\mathcal{R}}} = 0.249\) & \(\text{IG}_{\mathcal{T}_{\mathcal{R}}} = 0.122\) \\
        Demographics (\(\mathcal{D}\)) & \ 0.187 &\(\text{IG}_{\mathcal{D}} = 0.237\) & \(\text{IG}_{\mathcal{D}} = 0.250\) \\
        Images (\(\mathcal{I}\)) & \ 0.172 &\(\text{IG}_{\mathcal{I}} = 0.184\) & \(\text{IG}_{\mathcal{I}} = 0.207\) \\
        Time-Series (\(\mathcal{TS}\)) & \ 0.100 &\(\text{IG}_{\mathcal{TS}} = 0.066\) & \(\text{IG}_{\mathcal{TS}} = 0.086\) \\
        \bottomrule
    \end{tabular}
\end{table}

Following our performance analysis, we assess modality contributions in a task-specific supervised context to validate the \textit{modality-$\lambda$} used in mLSTM. Integrated Gradients (IG) estimates input importance by comparing a model’s output for a given input to a baseline. Unlike contrastive loss weights, which reflect modality importance during pre-training, IG reveals which modalities contribute most to final task predictions. This helps validate whether contrastively learned weights align with supervised behavior and offers a complementary view on modality relevance. Given a model, input, and baseline vectors (typically zero), IG estimates the importance of input features to the model’s output \cite{sundararajan2017axiomatic}. Using concatenated embeddings as input to the classifier (Figure \ref{fig:pipeline_overview} A.2), we calculated aggregated modality attribution scores for phenotype and IHM prediction.

As shown in Table \ref{tab:ig_scores}, the highest attribution scores across both tasks were seen in discharge text embeddings. This finding is consistent with the results in Table \ref{tab:combined_results}, where models incorporating discharge text achieve the highest performance across contrastive and fully-supervised settings. Conversely, time-series embeddings showed much lower attribution scores than other modalities. This aligns with our observation in Figure \ref{fig:CL_tsne} that time-series data is particularly challenging to integrate with other modalities. Overall, the attribution scores align with the contrastively learned modality lambdas in Table \ref{tab:ig_scores}.

\section{Discussion}
\subsection{Reproducibility and Benchmarking Considerations}

Our predictive tasks align with prior multimodal MIMIC studies \cite{hayat2022medfuse, liu2023attention, soenksen2022integrated}. A key distinction is our evaluation across 10 random seeds, as most existing studies report results from only a few seeds or none at all. Given the sensitivity of medical deep learning to model initialization, such variability can substantially affect reported outcomes. We advocate for greater transparency in multimodal MIMIC benchmarking and encourage reporting across multiple seeds, full performance distributions, and standardized evaluation practices.

\subsection{Limitations and Future Work}
Our study systematically evaluates multimodal learning by training on all MIMIC modality combinations in the unsupervised setting and the top three most effective in the supervised setting. This approach requires substantial compute (Appendix~\ref{app:training_compute}), limiting exploration of additional model architectures. To focus on contrastive loss and modality importance, we use encoders from prior work \cite{hayat2022medfuse, soenksen2022integrated, liu2023attention}. However, it is possible that alternative or more specialized encoders could yield richer representations, particularly for time-series data, which exhibited weaker alignment in the contrastive setting.

For consistency, we analyze only patients with complete data across all five modalities. This enables controlled evaluation but may not reflect real-world scenarios with missing modalities. Future work will extend the pipeline to handle missing data during inference. Contrastive pre-training may mitigate missing modality effects through shared representations, an avenue we plan to explore for improving robustness.

\section{Conclusions}
We introduce the Pipeline for Contrastive Modality Evaluation and Encoding (PiCME), the first to assess all 26 combinations of two to five modalities in the MIMIC dataset. Through our analysis of modality roles in prediction tasks, we identify strengths and limitations of contrastive pre-training for multimodal clinical models. Contrastive learning performs well with few modalities, offering reusable representations for multiple tasks. Performance peaks with three modalities: discharge summaries, chest X-rays, and demographics. However, performance plateaus or declines with more inputs, motivating our Modality-Gated LSTM, which dynamically adjusts modality contributions using contrastive weights. This approach yields the highest results in five-modality settings, improving predictive performance and generalization. We validate modality importance with attribution methods, showing gains in interpretability and fairness. Taken together, PiCME offers a robust and scalable pipeline for multimodal integration in clinical deep learning, and offers guidance for modality selection, training strategies, and equitable model design.

\section{Acknowledgments} 
We thank the MIMIC database (\url{https://physionet.org/content/mimiciv/3.1/}) and MIMIC-CXR (\url{https://physionet.org/content/mimic-cxr/2.1.0/}) for providing data for this study.

\printbibliography

\newpage 
\section{Appendix}
\label{section:supp-materials}

\subsection{Additional Details on Predictive Tasks}
We evaluate our approach on two clinical tasks: \textbf{in-hospital mortality} prediction (IHM) and \textbf{phenotyping}. The IHM prediction task is a binary classification problem that aims to predict whether a patient will survive their hospital stay based on data from the first 48 hours of an ICU admission. We exclude ICU stays shorter than 48 hours to ensure a consistent observation period (following \textcite{hayat2022medfuse}). Due to the significant class imbalance in this task, we employ a weighted binary cross-entropy loss, where the loss is scaled according to the number of patients in each class to prevent the model from being biased toward the majority class.

The phenotype classification task is a multi-label classification problem, where the goal is to predict whether a patient has any of 25 chronic, mixed, or acute care conditions during an ICU stay. Phenotype labels are derived from ICD-9 and ICD-10 codes, mapped to Clinical Classifications Software (CCS) categories. Since each patient can have multiple conditions, we apply a categorical cross-entropy loss without weighting, allowing independent optimization for each label.

\subsection{Hyperparameters}\label{app:exp_set}

For both unsupervised and supervised training, we tuned learning rates ranging from \(10^{-2}\) to \(10^{-6}\) and batch sizes from \(16\) to \(128\). We set the maximum number of epochs to 75 and use early stopping if the validation
AUROC does not improve for 15 epochs. All models were first evaluated on the validation set using a single seed to determine the best hyperparameters. Only after selecting the optimal configuration, the final model was run on the test set across 10 random seeds, with results reported as the average performance.

\subsection{Data Splits}
We use the dataset splits from \cite{hayat2022medfuse} and further filter for patients with all five modalities present. This results in X training samples for in-hospital mortality and X for phenotyping. Contrastive learning uses the full original training set, while downstream fine-tuning is performed on a 80/10/10 split of the remaining data. Final evaluation is reported on the fine-tuning test split.

\subsection{Demographics counts}
The following table presents the sample counts for each demographic group in the test set, providing important context for interpreting performance variability.

\begin{table}[ht]
\centering
\begin{tabular}{lll}
\toprule
\textbf{Group} & \textbf{Subgroup} & \textbf{Count} \\
\midrule
\multirow{2}{*}{Gender} 
  & Female & 866 \\
  & Male   & 1164 \\
\midrule
\multirow{5}{*}{Ethnicity} 
  & Asian                     & 55 \\
  & Black / African American & 333 \\
  & Hispanic / Latino        & 51 \\
  & Unknown                  & 355 \\
  & White                    & 1236 \\
\midrule
\multirow{8}{*}{Age Group} 
  & Under 25 & 10 \\
  & 25–34    & 78 \\
  & 35–44    & 49 \\
  & 45–54    & 528 \\
  & 55–64    & 528 \\
  & 65–74    & 357 \\
  & 75–84    & 242 \\
  & 85+      & 238 \\
\bottomrule
\end{tabular}
\caption{Test set distribution across gender, ethnicity, and age group. Small subgroup sizes (e.g., Under 25, Asian, Latino) may contribute to higher performance variance.}
\label{tab:demo_distribution}
\end{table}

\subsection{Performance by Phenotype Type}

To evaluate how training strategies perform across different clinical label types, we group the 25 phenotype labels into broader categories: \textit{acute}, \textit{chronic}, and \textit{mixed} \cite{medfuse}. Table~\ref{tab:pheno_type_results} reports AUROC scores for each group using all five modalities, comparing contrastive pre-training, fully fine-tuned models, and the Modality-Gated LSTM (mLSTM).

\begin{table}[ht]
\centering
\begin{tabular}{lcc}
\toprule
\textbf{Phenotype Type} & \textbf{Training Strategy} & \textbf{Mean AUROC ± Std.} \\
\midrule
\multirow{3}{*}{Acute} 
  & Contrastive            & 67.40 ± 0.98 \\
  & Supervised Baseline        & 69.50 ± 0.11 \\
  & \textbf{Modality-Gated LSTM} & \textbf{71.30 ± 0.50} \\
\midrule
\multirow{3}{*}{Chronic} 
  & Contrastive            & 67.60 ± 0.50 \\
  & Supervised Baseline        & 76.50 ± 0.42 \\
  & \textbf{Modality-Gated LSTM} & \textbf{76.80 ± 0.12} \\
\midrule
\multirow{3}{*}{Mixed} 
  & Contrastive            & 71.90 ± 0.08 \\
  & Supervised Baseline        & 77.20 ± 0.04 \\
  & \textbf{Modality-Gated LSTM} & \textbf{77.70 ± 0.04} \\
\bottomrule
\end{tabular}
\caption{Phenotyping AUROC performance across phenotype types. Values are reported as mean ± standard deviation (\%).}
\label{tab:pheno_type_results}
\end{table}

Across all phenotype types (acute, chronic, and mixed), the Modality-Gated LSTM achieves the highest AUROC, providing the most performance gains in the acute category where overall performance is lowest. Acute phenotypes represent rapidly emerging conditions like sepsis or respiratory failure, which are harder to detect due to their short time window, rapidly shifting clinical signals, and increased noise across modalities, particularly in time-series and unstructured text. When integrating five modalities, this noise is amplified, and fixed-fusion approaches struggle to isolate relevant information. Contrastive learning, which relies on frozen, task-agnostic representations, is especially vulnerable in this setting, as it cannot adaptively down-weight noisy or uninformative signals. Chronic conditions, by contrast, unfold more gradually and consistently, allowing even simpler models to identify stable cross-modal patterns. Mixed phenotypes combine both types and achieve the highest performance overall, likely because chronic context provides a stable baseline while acute events introduce sharper, complementary features. In these complex cases, the Modality-Gated LSTM’s ability to dynamically reweight modalities proves particularly effective. These findings highlight how structured fusion mechanisms are critical in navigating the complexity and signal imbalance inherent in real-world clinical data.

\subsection{Training Infrastructure}\label{app:training_compute}
All experiments in this project use one NVIDIA GeForce RYX 3090 GPU. For contrastive pre-training, these jobs ran from 3 to 10 hours with early-stopping implemented as a function of model performance on the validation dataset. In comparison, training fine-tuning heads on top of the contrastively learned weights ran for roughly 2 and 3 hours for IHM and phenotyping, respectively. \

\begin{table}[ht]
\centering
\begin{tabular}{llcc}
\toprule
\textbf{Modalities} & \textbf{Task} & \textbf{Concatenation} & \textbf{mLSTM} \\
\midrule
\multirow{2}{*}{3} 
  & IHM          & 3 hrs & 2.2 hrs\\
  & Phenotyping  & 6.5 hrs & 4.5 hrs \\
\midrule
\multirow{2}{*}{4} 
  & IHM          & hrs & hrs \\
  & Phenotyping  &  hrs &  hrs \\
\midrule
\multirow{2}{*}{5} 
  & IHM          & hrs & hrs \\
  & Phenotyping  &  hrs &  hrs \\
\bottomrule
\end{tabular}
\caption{Approximated runtime for mLSTM and concatenation in fully finetuned baseline models across the IHM and phenotyping tasks.}
\end{table}

In the fully fine-tuned baseline setting, we can examine the difference in runtime between the concatenation and Modality-Gated LSTMs for both IHM and phenotyping. In Table 7, we see a pattern that fine-tuning with mLSTM tends to converge slightly faster than fine-tuning with concatenation runs. Across all architectures and modality combinations, we observe that training converges slower for phenotyping than for IHM. 

\end{document}